\documentclass[letterpaper]{article} 
\usepackage{aaai2026}  
\usepackage{times}  
\usepackage{helvet}  
\usepackage{courier}  
\usepackage[hyphens]{url}  
\usepackage{graphicx} 
\usepackage{natbib}  
\usepackage{caption} 
\usepackage{algorithm}
\usepackage{algorithmic}

\usepackage{amssymb}
\usepackage{amsmath}
\usepackage{multirow}
\usepackage{booktabs}
\usepackage{array}

\usepackage{newfloat}
\usepackage{listings}
\DeclareCaptionStyle{ruled}{labelfont=normalfont,labelsep=colon,strut=off} 
\floatstyle{ruled}
\newfloat{listing}{tb}{lst}{}
\floatname{listing}{Listing}
\title{Bridging Granularity Gaps: Hierarchical Semantic Learning for Cross-domain Few-shot Segmentation}
\author{
    Sujun Sun\textsuperscript{\rm 1, \rm 2},
    Haowen Gu\textsuperscript{\rm 1},
    Cheng Xie\textsuperscript{\rm 1},
    Yanxu Ren\textsuperscript{\rm 1},
    Mingwu Ren\textsuperscript{\rm 1, \rm 2},
    Haofeng Zhang\textsuperscript{\rm 1, \rm 2, }\thanks{Corresponding Author.}\\
}
\affiliations{
    \textsuperscript{\rm 1}School of Computer Science and Engineering, Nanjing University of Science and Technology, China\\
    \textsuperscript{\rm 2}State Key Laboratory of Intelligent Manufacturing of Advanced Construction Machinery, China\\

    \{egg, guhaowen, city945, 324106010241, renmingwu, zhanghf\}@njust.edu.cn

}

\usepackage{bibentry}

\begin{document}

\maketitle

\begin{abstract}
Cross-domain Few-shot Segmentation (CD-FSS) aims to segment novel classes from target domains that are not involved in training and have significantly different data distributions from the source domain, using only a few annotated samples, and recent years have witnessed significant progress on this task. However, existing CD-FSS methods primarily focus on style gaps between source and target domains while ignoring segmentation granularity gaps, resulting in insufficient semantic discriminability for novel classes in target domains. Therefore, we propose a Hierarchical Semantic Learning (HSL) framework to tackle this problem. Specifically, we introduce a Dual Style Randomization (DSR) module and a Hierarchical Semantic Mining (HSM) module to learn hierarchical semantic features, thereby enhancing the model's ability to recognize semantics at varying granularities. DSR simulates target domain data with diverse foreground-background style differences and overall style variations through foreground and global style randomization respectively, while HSM leverages multi-scale superpixels to guide the model to mine intra-class consistency and inter-class distinction at different granularities. Additionally, we also propose a Prototype Confidence-modulated Thresholding (PCMT) module to mitigate segmentation ambiguity when foreground and background are excessively similar. Extensive experiments are conducted on four popular target domain datasets, and the results demonstrate that our method achieves state-of-the-art performance. Our code is available at \url{https://github.com/Sparkling-Water/HSLNet}.
\end{abstract}


\begin{figure}[t]
    \centering
    \includegraphics[width=0.99\linewidth]{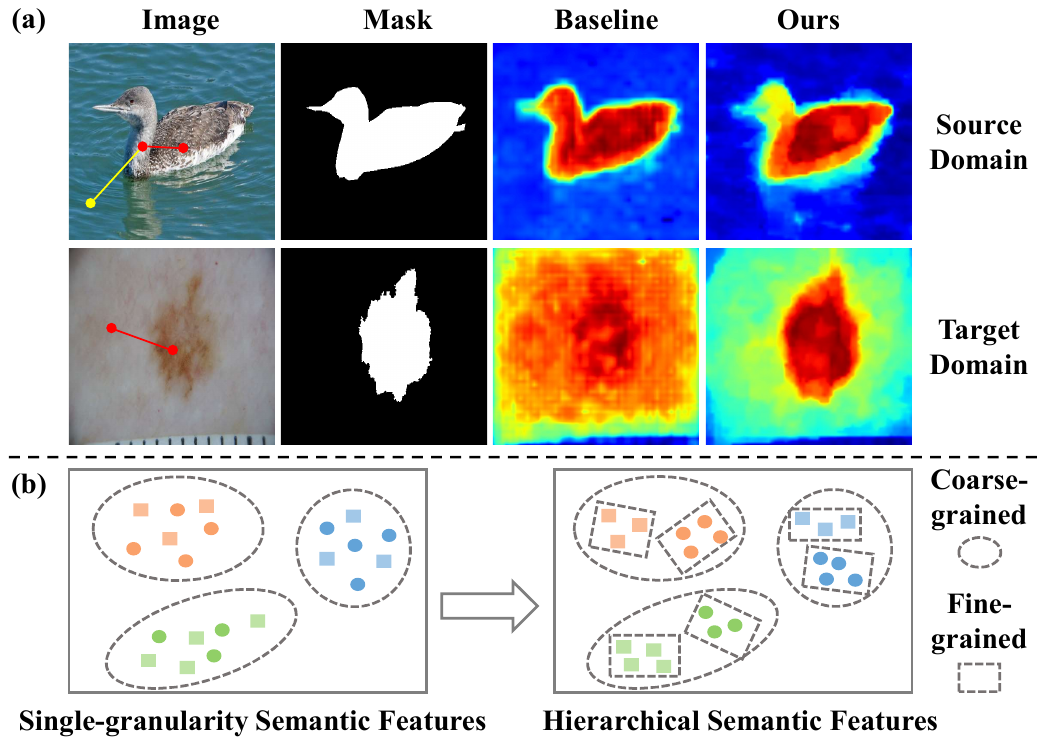}
    \caption{Motivation of the proposed method. (a) Segmentation granularity gaps between the source and target domains, \textit{e.g.}, the foreground-background differences in the target domain are more similar to the finer-grained differences within the foreground of the source domain. The model trained on the source domain only focuses on single-granularity differences between training classes and background, with insufficient ability to distinguish the foreground and background in the target domain. (b) We aim to extract hierarchical semantic features to adapt to target domain data with different segmentation granularities.}
    \label{fig:motivation}
\end{figure}

\section{Introduction}

In recent years, deep learning has achieved remarkable progress in numerous computer vision tasks \cite{hu2023you, wang2024yolov10, bai2024artrackv2}. However, this success is typically predicated on large-scale annotated datasets \cite{yu2020bdd100k, liao2022kitti, huang2019got}, whose construction is time-consuming and labor-intensive, particularly for dense prediction tasks like semantic segmentation. To overcome data scarcity, Few-shot Segmentation (FSS) has been proposed and widely studied \cite{peng2023hierarchical, sun2024vrp, xu2024eliminating}, and it learns generalizable category correspondence via meta-learning on base classes with abundant annotated samples, thereby enabling the segmentation of novel classes using only a few annotated samples. Nevertheless, existing FSS methods assume base classes and novel classes originate from the same domain. In cross-domain scenarios, these methods suffer significant performance degradation due to domain gaps between base classes and novel classes, severely limiting their practical applications.

To tackle both few-shot and cross-domain challenges simultaneously, Cross-domain Few-shot Segmentation (CD-FSS) \cite{lei2022cross} has recently gained wide attention. Existing CD-FSS methods typically mitigate overfitting to source domain by applying style perturbation to features \cite{su2024domain, liu2025devil}, transforming features into a domain-agnostic spaces \cite{lei2022cross, he2024apseg}, or fine-tuning model parameters using limited target domain annotated data \cite{nie2024cross, peng2025sam, herzog2024adapt}. Since fine-tuning with target domain images significantly increases computational costs during testing and easily disrupts cross-view semantic consistency, we focus on methods without target domain fine-tuning. However, we observe that these methods primarily address inter-domain style variations but ignore a core issue: segmentation granularity gaps exist between source and target domains, limiting their performance improvement.

Specifically, as shown in Fig. \ref{fig:motivation}(a), the foreground (bird) and background (other classes) in the source domain image exhibit distinct differences, which we refer to as a coarse-grained semantic partition. In contrast, the distinction between the bird's neck and wings, or between its brown and gray feathers, is an example of a fine-grained semantic partition. Since the model is trained only on source domain with the aim of reducing intra-class differences and increase inter-class differences for training classes, it tends to extract homogeneous foreground features and treats the entire foreground object as a single entity, ignoring fine-grained differences within the foreground. The foreground (skin lesion area) and background (normal skin area) in the target domain image show less contrast, resembling the fine-grained differences of bird feather colors. Consequently, the model trained on the source domain struggles to effectively capture the finer-grained foreground and background semantic partitions in target domains. To solve this problem, an intuitive idea is to extract hierarchical semantic features that exhibit intra-class consistency and inter-class distinction at different granularities, as shown in Fig. \ref{fig:motivation}(b), thereby ensuring the model better adapt to granularity gaps.

Based on the above analysis, we propose a Hierarchical Semantic Learning (HSL) framework for CD-FSS, which consists of three key modules. First, we propose a Dual Style Randomization (DSR) module that sequentially performs foreground style randomization and global style randomization on training data. Specifically, foreground style randomization modulates the foreground style by using randomly selected local region styles, enhancing the model's ability to capture varying degrees of inter-class differences without destroying the semantic content of images. Global style randomization modulates the overall image style via random convolution, adapting the model to diverse inter-domain style variations. To further enhance the hierarchical partitioning ability of original features, we propose a Hierarchical Semantic Mining (HSM) module. Since multi-scale superpixel segmentation masks naturally partition an image into local regions approximating semantic regions at different granularities, HSM leverages these masks to generate multi-scale region prototypes for both low-level and high-level features. This process explicitly enhances the inter-class distinction while implicitly guiding the model to mine intra-class consistency across different granularities. Finally, observing that existing similarity comparison-based segmentation methods lead to segmentation ambiguity when foreground and background are excessively similar, we introduce a Prototype Confidence-modulated Thresholding (PCMT) module, which utilizes prototype confidence-weighted adaptive thresholds to segment foreground confidence maps, resulting in more accurate query masks. 
In summary, our contributions are as follows:
\begin{itemize}
\item We are the first to focus on segmentation granularity gaps between source and target domains in CD-FSS, and propose a DSR module and an HSM module to learn hierarchical semantic features, enabling the model to generalize better to target domains with varying segmentation granularities.
\item We propose a PCMT module to mitigate segmentation ambiguity when foreground and background are excessively similar, thereby improving segmentation accuracy.
\item Extensive experiments on four standard target domain datasets demonstrate that our method significantly outperforms current state-of-the-art CD-FSS methods.
\end{itemize}

\section{Related Works}

\subsection{Few-shot Segmentation}
FSS aims to segment novel classes in query images using only a few annotated images. Existing FSS methods can be broadly categorized into two types: prototype-based methods and matching-based methods. Prototype-based methods typically extract single \cite{zhang2019canet, fan2022self, sun2024vrp} or multiple prototypes \cite{yang2020prototype, liu2020part, li2021adaptive} from support images as category guidance and use feature concatenation or distance calculation to segment query images. Due to spatial structure information loss during prototype computation, numerous matching-based methods \cite{peng2023hierarchical, xu2023self, xu2024eliminating} have been proposed to fully exploit information in support images. These methods focus on calculating pixel-to-pixel dense correlations between support and query features, decoding them to obtain query masks. However, when base classes and novel classes originate from source and target domains with significant domain gaps, these FSS methods often fail to generalize effectively to novel classes, resulting in a sharp decline in segmentation performance.

\begin{figure*}[t]
\centering
    \includegraphics[width=0.97\linewidth]{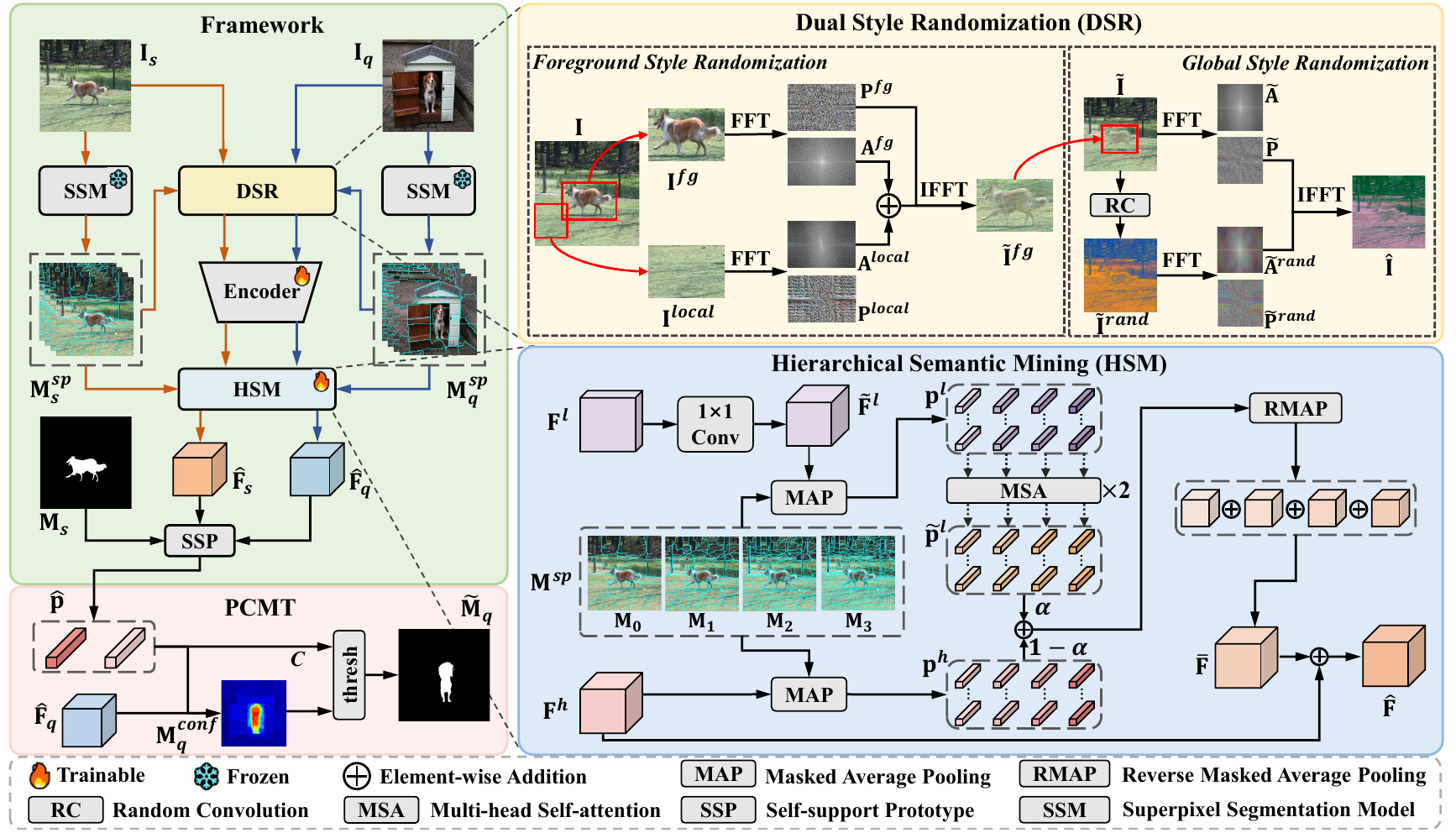}
    \caption{Overview of our method. We first use the SSM to extract multi-scale superpixel masks for both support and query images. These masks are applied to the DSR and HSM to assist in extracting hierarchical semantic features. Subsequently, support and query images are sequentially fed into the DSR, image encoder, and HSM for data augmentation, feature extraction, and feature enhancement, resulting in hierarchical semantic features. Finally, we compute foreground and background prototypes through SSP. These prototypes and query features are fed into the PCMT to perform query image segmentation.}
    \label{fig:overall architecture}
\end{figure*}

\subsection{Cross-domain Few-shot Segmentation}
To overcome the limitations of FSS methods, many CD-FSS methods have been proposed, aiming to ensure that the model trained solely on the source domain can generalize to target domains with large domain gaps. Some methods \cite{su2024domain, liu2025devil} introduce style perturbations during training to simulate style variations across domains, extracting features with better generalization. Other methods design complex support-query matching mechanisms to better exploit the correlations between support and query images, though they often require additional strategies to mitigate overfitting, such as transforming features extracted by the backbone into domain-agnostic spaces \cite{lei2022cross, he2024apseg} or fine-tuning with a few annotated samples from target domains \cite{nie2024cross, tong2024lightweight, peng2025sam, tong2025self}. Some works \cite{herzog2024adapt} completely abandon the training stage and directly fine-tune with limited annotated data. In contrast, we focus on extracting hierarchical semantic features that exhibit intra-class consistency and inter-class distinction at different granularities, without the need for complicated fine-tuning processes, while simultaneously adapting to inter-domain style variations and segmentation granularity differences.

\section{Methodology}

\subsection{Problem Definition}
In the CD-FSS task, we have two distinct data domains: a source domain $\mathcal{D}_s = \{X_s, Y_s\}$ and a target domain $\mathcal{D}_t = \{X_t, Y_t\}$, where $X_s$ and $X_t$ represent data distributions, and $Y_s$ and $Y_t$ denote corresponding label spaces. The task satisfies $X_s \neq X_t$ and $Y_s \cap Y_t = \emptyset$, indicating no overlap in both data distributions and class labels.

The CD-FSS framework follows a meta-learning paradigm, where only source domain data $\mathcal{D}_{train}\subseteq \mathcal{D}_s$ is used during training, while the model's generalization ability to novel classes is evaluated on $\mathcal{D}_{test}\subseteq \mathcal{D}_t$ during testing. Each training or testing task is organized as an episode, containing a support set $\mathcal{S}=\{(\mathbf{I}^i_s,\mathbf{M}^i_s)\}_{i=1}^{K}$ and a query set $\mathcal{Q}=\{(\mathbf{I}_q,\mathbf{M}_q)\}$, where $K$ denotes the number of support samples, $\mathbf{I}_s$ and $\mathbf{I}_q$ represent support and query images, and $\mathbf{M}_s$ and $\mathbf{M}_q$ are their corresponding binary masks. In each episode, the model predicts the query mask based on the support set $\mathcal{S}$ and the query image $\mathbf{I}_q$.

\subsection{Method Overview}
The architecture of our method is illustrated in Fig. \ref{fig:overall architecture}, which consists of three key modules to address style gaps and granularity gaps: a Dual Style Randomization (DSR) module, a Hierarchical Semantic Mining (HSM) module, and a Prototype Confidence-modulated Thresholding (PCMT) module.

First, multi-scale superpixel masks for support and query images are obtained through a superpixel segmentation model \cite{xu2024learning}. These masks are applied to the DSR and HSM modules to assist in extracting hierarchical semantic features. Next, DSR sequentially performs foreground and global style randomization on the input images. The randomized images are then fed into an image encoder to extract their corresponding features, which are further enhanced in their hierarchical discrimination ability through HSM. Subsequently, the enhanced features and support mask are input into the SSP module \cite{fan2022self} to extract foreground and background prototypes. Finally, PCMT utilizes these prototypes and the query feature to compute the foreground confidence map, which is segmented using a prototype confidence-weighted adaptive threshold to generate the final query prediction. We use binary cross entropy (BCE) loss to train our model, as detailed in the appendix. Note that DSR is used only during training, while PCMT is employed exclusively during testing.

\subsection{Dual Style Randomization}
To mitigate performance degradation caused by domain gaps, prior works \cite{su2024domain, liu2025devil} randomize the global style of input images or features to simulate target domain data with diverse styles. Building upon this, we propose a DSR module, which introduces an additional foreground style randomization process to enhance the model's ability to capture varying degrees of inter-class differences.

\subsubsection{Foreground Style Randomization.} 
We aim to simulate target domain data with varying inter-class differences, allowing the model to adapt to segmentation granularity gaps. To achieve this, we use styles from randomly selected local regions to modulate the foreground style without destroying the semantic content of images.

First, we extract the foreground image and the local region image. Given an original image $\mathbf{I}\in \mathbb{R}^{3\times H\times W}$ and its corresponding foreground mask $\mathbf{M}\in\mathbb{R}^{H\times W}$, where $H$ and $W$ represent the image height and width, we preprocess $\mathbf{M}$ via max pooling and average pooling with kernel size $K\times K$ to obtain the mask $\mathbf{M}^{\prime}$. Subsequently, we compute the bounding box of the foreground from $\mathbf{M}^{\prime}$ and crop the corresponding region from $\mathbf{I}$, thus obtaining the foreground region image $\mathbf{I}^{fg}$. Simultaneously, given multi-scale superpixel masks $\mathcal{M}^{sp}=\{\mathbf{M}_i\}_{i=1}^L$ for $\mathbf{I}$, where $L$ represents the number of superpixel scales, $\mathbf{M}_i\in\mathbb{R}^{H\times W}$ divides the image into $n_i$ superpixel regions $\{R_{i1},R_{i2},...,R_{in_i}\}$, and $n_i<n_{i+1}$. We randomly select the $j$-th superpixel region $R_{0j}$ from the coarsest superpixel mask $\mathbf{M}_0$. By computing the bounding box corresponding to $R_{0j}$ and cropping the corresponding region from $\mathbf{I}$, we obtain the randomly selected local region image $\mathbf{I}^{local}$, which is then resized to the same size as $\mathbf{I}^{fg}$.

The Fast Fourier Transformation (FFT) can convert images from the spatial domain to the frequency domain and further decompose them into style-representing amplitude spectrum and content-representing phase spectrum. Some works \cite{chen2021amplitude, liu2025devil} achieve style transformation of images or features by retaining the phase and replacing the amplitude. Here, we compute the random weighted sum of the amplitude of $\mathbf{I}^{fg}$ and $\mathbf{I}^{local}$, allowing the foreground region’s style to approach or diverge from the local region’s style to varying degrees. Specifically, we transform $\mathbf{I}^{fg}$ and $\mathbf{I}^{local}$ to the frequency domain via FFT and decompose them into phase spectrum $\mathbf{P}^{fg}$, $\mathbf{P}^{local}$ and amplitude spectrum $\mathbf{A}^{fg}$, $\mathbf{A}^{local}$:
\begin{align}
\label{eq:fft}
[\mathbf{A}^{fg}, \mathbf{P}^{fg}]&=\operatorname{FFT}(\mathbf{I}^{fg}),\\
[\mathbf{A}^{local}, \mathbf{P}^{local}]&=\operatorname{FFT}(\mathbf{I}^{local}).
\end{align}

Next, we obtain the amplitude spectrum $\mathbf{A}^{fusion}$ representing the randomized foreground style by randomly weighting $\mathbf{A}^{fg}$ and $\mathbf{A}^{local}$:
\begin{equation}
\mathbf{A}^{fusion}=\omega\mathbf{A}^{local}+(1-\omega)\mathbf{A}^{fg},
\end{equation}
where the weight parameter $\omega$ is sampled from a normal distribution $N(0,\sigma_{f}^2)$. We recombine $\mathbf{P}^{fg}$ and $\mathbf{A}^{fusion}$ through the Inverse Fast Fourier Transform (IFFT), generating the style-randomized foreground region image $\tilde{\mathbf{I}}^{fg}$:
\begin{equation}
\label{eq:ifft}
\tilde{\mathbf{I}}^{fg}=\operatorname{IFFT}(\mathbf{A}^{fusion}e^{i\mathbf{P}^{fg}}),
\end{equation}
and further pad $\tilde{\mathbf{I}}^{fg}$ to the same size as the original image $\mathbf{I}$.

Finally, we use $\tilde{\mathbf{I}}^{fg}$ to replace the foreground region of $\mathbf{I}$:
\begin{equation}
\tilde{\mathbf{I}}=\mathbf{M}^{\prime}\mathbf{I}+(1-\mathbf{M}^{\prime})\tilde{\mathbf{I}}^{fg},
\end{equation}
where $\tilde{\mathbf{I}}$ is the foreground style-randomized image. Since the mask $\mathbf{M}^{\prime}$ has been preprocessed with max pooling and average pooling, it ensures a smooth transition between the foreground and background regions, preventing obvious foreground-background edges in $\tilde{\mathbf{I}}$.

\subsubsection{Global Style Randomization.}
Global style randomization aims to perturb the overall style of each training image, enabling the model to adapt to style variations across domains. Our global style randomization is similar to \cite{liu2025devil}, but we apply it to image $\tilde{\mathbf{I}}$ instead of the middle-layer features. We first input $\tilde{\mathbf{I}}$ into a random convolution (RC) layer, generating an image $\tilde{\mathbf{I}}^{rand}$ with random style by perturbing the local textures of $\tilde{\mathbf{I}}$:
\begin{equation}
\tilde{\mathbf{I}}^{rand}=\operatorname{RC}(\tilde{\mathbf{I}}),
\end{equation}
where the parameters $\mathbf{\Theta} \in\mathbb{R}^{3\times3\times3\times3}$ of the convolution kernel are sampled from a normal distribution $N(0,\sigma_{g}^2)$.

Although random convolution can randomize the overall style of an image, it inevitably destroys some content details. Therefore, instead of directly using $\tilde{\mathbf{I}}^{rand}$ for training, we use the style of $\tilde{\mathbf{I}}^{rand}$ to modulate the style of $\tilde{\mathbf{I}}$. Specifically, following Eq. \ref{eq:fft}, we convert $\tilde{\mathbf{I}}$ and $\tilde{\mathbf{I}}^{rand}$ to the frequency domain via FFT and decompose them into phase spectrum $\tilde{\mathbf{P}}$, $\tilde{\mathbf{P}}^{rand}$ and amplitude spectrum $\tilde{\mathbf{A}}$, $\tilde{\mathbf{A}}^{rand}$. Then, following Eq. \ref{eq:ifft}, we directly recombine $\tilde{\mathbf{P}}$ and $\tilde{\mathbf{A}}^{rand}$ via IFFT to generate the style-randomized image $\hat{\mathbf{I}}$.

\subsection{Hierarchical Semantic Mining}
In addition to performing data augmentation in DSR, we also process the features to further enhance their hierarchical discrimination ability. We observe that superpixel segmentation \cite{yang2020superpixel, wang2021ainet, xu2024learning} can divide an image into multiple local regions, each consisting of a set of adjacent pixels with similar features or semantics. By introducing superpixel masks at multiple scales, we can approximate semantic regions at different granularities, thereby guiding the model to mine intra-class consistency and inter-class distinction at different granularities.

Specifically, given multi-scale superpixel masks $\mathcal{M}^{sp}=\{\mathbf{M}_i\}_{i=1}^L$ corresponding to the original image $\mathbf{I}$, we first generate binary masks $\{\tilde{\mathbf{M}}_{i1},\tilde{\mathbf{M}}_{i2},...,\tilde{\mathbf{M}}_{in_i}\}$ for the $n_i$ superpixel regions corresponding to $\mathbf{M}_i$, where the calculation process for the mask $\tilde{\mathbf{M}}_{ij}$ of the $j$-th region is formulated as:
\begin{equation}
\tilde{\mathbf{M}}_{ij}(x,y)=\begin{cases}1, \quad \mathbf{M}_i(x,y)=j  \\0, \quad \mathrm{otherwise}\end{cases},
\end{equation}
where $(x,y)$ is the spatial coordinate of the mask. Concurrently, we use an image encoder to extract the shallow low-level feature $\mathbf{F}^l \in \mathbb{R}^{c_l\times h_l\times w_l}$ and deep high-level feature $\mathbf{F}^h \in \mathbb{R}^{c_h\times h_h\times w_h}$, where $c_{*}$, $h_{*}$ and $w_{*}$ denote the channel depth, height and width of the feature, respectively.

Next, we compute low-level and high-level prototypes corresponding to multi-scale regions and leverage these prototypes to enhance features. High-level region prototypes encode semantics, while low-level ones contain details such as color and texture to enhance the distinction of features. Specifically, we first input $\mathbf{F}^l$ into a $1\times1$ convolution layer and downsample it to match the dimensions of $\mathbf{F}^h$:
\begin{equation}
\tilde{\mathbf{F}}^{l}=\operatorname{Down}(\operatorname{Conv}(\mathbf{F}^l))\in \mathbb{R}^{{c_h\times h_h\times w_h}}.
\end{equation}
Then, we use masked average pooling (MAP) to generate prototypes for each region:
\begin{align}
\mathbf{p}^{l}_{ij}&=\operatorname{MAP}(\tilde{\mathbf{F}}^l,\tilde{\mathbf{M}}_{ij}),\\
\mathbf{p}^{h}_{ij}&=\operatorname{MAP}(\mathbf{F}^h,\tilde{\mathbf{M}}_{ij}),
\end{align}
where $\mathbf{p}^{l}_{ij},\mathbf{p}^{h}_{ij}\in \mathbb{R}^{c_h}$ are the low-level and high-level region prototypes corresponding to the $j$-th region of the $i$-th scale superpixel mask, respectively.

Since low-level features typically contain noise and have a significant semantic gap with high-level features, we sequentially enhancing low-level prototypes at each scale through two multi-head self-attention (MSA) layers:
\begin{equation}
\tilde{\mathcal{P}}^{l}_{i}=\operatorname{MSA}_2(\operatorname{MSA}_1(\mathcal{P}^{l}_{i})),
\end{equation}
where $\mathcal{P}^{l}_{i}=\{\mathbf{p}^{l}_{ij}\}_{j=1}^{n_i}$ and $\tilde{\mathcal{P}}^{l}_{i}=\{\tilde{\mathbf{p}}^{l}_{ij}\}_{j=1}^{n_i}$ are the sets of all low-level prototypes at the $i$-th scale. Subsequently, we perform weighted fusion of $\tilde{\mathbf{p}}^{l}_{ij}$ and $\mathbf{p}^{h}_{ij}$:
\begin{equation}
\mathbf{p}_{ij}=\alpha \tilde{\mathbf{p}}^{l}_{ij}+(1-\alpha)\mathbf{p}^{h}_{ij},
\end{equation}
where $\alpha$ is a weight parameter.

To apply each region prototype to its corresponding pixels, we use the inverse process of MAP, namely RMAP \cite{peng2025sam}, to restore region prototypes for each scale into feature maps. This process is formulated as:
\begin{equation}
\bar{\mathbf{F}}_i(x,y)=\sum^{n_i}_{j}\mathbf{p}_{ij}\tilde{\mathbf{M}}_{ij}(x,y),
\end{equation}
where $\bar{\mathbf{F}}_i \in \mathbb{R}^{c_h\times h_h\times w_h}$ is the region feature map at scale $i$. Finally, we use region feature maps at all scales to enhance $\mathbf{F}^h$, thereby obtaining hierarchical semantic features $\hat{\mathbf{F}}$:
\begin{equation}
\hat{\mathbf{F}}=\mathbf{F}^h+\sum_{i}^{L}\bar{\mathbf{F}}_i.
\end{equation}

Since each pixel's segmentation is influenced by multi-scale region prototypes, this guides the model to focus on relationships between multi-scale regions across views during training, thereby enhancing the intra-class consistency and inter-class distinction of $\hat{\mathbf{F}}$ at different granularities.

\subsection{Prototype Confidence-modulated Thresholding}
After obtaining the hierarchical semantic features $\hat{\mathbf{F}}_s$ and $\hat{\mathbf{F}}_q$, we compute the support foreground prototype $\hat{\mathbf{p}}^{fg}_s$ and background prototype $\hat{\mathbf{p}}^{bg}_s$ through MAP, further using the SSP module \cite{fan2022self} to compute the query prototypes $\hat{\mathbf{p}}^{fg}_q$ and $\hat{\mathbf{p}}^{bg}_q$, as well as the fused prototypes $\hat{\mathbf{p}}^{fg}$ and $\hat{\mathbf{p}}^{bg}$. Subsequently, we upsample the query feature $\hat{\mathbf{F}}_q$ to the size of $\mathbf{I}_q$, computing the cosine similarity between the fused prototypes and $\hat{\mathbf{F}}_q$ to obtain a foreground similarity map $\mathbf{M}^{fg}_q=\operatorname{cos}(\hat{\mathbf{p}}^{fg},\hat{\mathbf{F}}_q)\in\mathbb{R}^{H\times W}$ and a background similarity map $\mathbf{M}^{bg}_q=\operatorname{cos}(\hat{\mathbf{p}}^{bg},\hat{\mathbf{F}}_q)\in\mathbb{R}^{H\times W}$.

Previous works \cite{fan2022self, su2024domain, wang2019panet} obtain final query predictions based on similarity comparison, directly classifying pixels as foreground where foreground similarity exceeds background similarity. We found that this approach easily leads to segmentation ambiguity in cross-domain scenarios. Specifically, for images with excessively similar foreground and background, the distinction of their foreground and background features is limited. However, due to cross-view intra-class variations, it is inevitable that both query foreground and background features are more similar to the foreground prototype (or background prototype). In such cases, the similarity comparison-based method results in incorrect segmentation.

Inspired by \cite{herzog2024adapt}, we introduce the idea of threshold-based segmentation, using prototype confidence-weighted adaptive thresholds to segment the foreground confidence maps. Specifically, we first compute the query foreground confidence map $\mathbf{M}^{conf}_q$:
\begin{equation}
\mathbf{M}^{conf}_q=\mathbf{M}^{fg}_q-\mathbf{M}^{bg}_q.
\end{equation}
Next, we compute the adaptive threshold $t$ via OTSU \cite{otsu1975threshold} to segment $\mathbf{M}^{conf}_q$ into a binary mask.

Experiments reveal that the adaptive threshold $t$ effectively mitigates segmentation ambiguity, but it introduces additional errors on data where no ambiguity exists. To address this, we introduce prototype confidence to modulate $t$. The prototype confidence $C$ is calculated as:
\begin{equation}
C=\operatorname{cos}(\hat{\mathbf{p}}^{fg}_{q}, \hat{\mathbf{p}}^{fg}_{s})-\frac{\operatorname{cos}(\hat{\mathbf{p}}^{fg}_{q}, \hat{\mathbf{p}}^{bg}_{s})+\operatorname{cos}(\hat{\mathbf{p}}^{fg}_{s}, \hat{\mathbf{p}}^{bg}_{q})}{2},
\end{equation}
which approximates the probability of segmentation ambiguity based on cross-view prototype similarity. We then segment $\mathbf{M}^{conf}_q$ using the adaptive threshold weighted by $C$:
\begin{equation}
\label{eq:segment}
\tilde{\mathbf{M}}_q=\mathbf{M}^{conf}_q >\frac{1}{1+e^{\beta(C+\gamma)}}t,
\end{equation}
where $\beta$ and $\gamma$ are hyperparameters that adjust the weight. 

In fact, when the threshold is fixed to 0, Eq. \ref{eq:segment} is equivalent to the similarity comparison-based method. For images with higher prototype confidence, we tend to use the threshold 0, whereas for images with lower prototype confidence, we prefer the adaptive threshold $t$.

\begin{table*}[t]
    \centering
    \setlength{\tabcolsep}{6pt}
    \resizebox{\textwidth}{!}{
    \begin{tabular}{l|c|c|cc|cc|cc|cc|cc}
        \toprule
        \rule{0pt}{2.5ex}
        \multirow{2}*{Methods} & \multirow{2}*{Backbone} & \multirow{2}*{Mark} & \multicolumn{2}{c|}{Deepglobe} & \multicolumn{2}{c|}{ISIC} & \multicolumn{2}{c|}{Chest X-ray} & \multicolumn{2}{c|}{FSS-1000} & \multicolumn{2}{c}{Average} \\
        \cline{4-13}
        \rule{0pt}{2.5ex}
        & & & \multicolumn{1}{c|}{1-shot} & 5-shot & \multicolumn{1}{c|}{1-shot} & 5-shot & \multicolumn{1}{c|}{1-shot} & 5-shot & \multicolumn{1}{c|}{1-shot} & 5-shot & \multicolumn{1}{c|}{1-shot} & 5-shot \\
        \midrule
        \multicolumn{13}{c}{Few-shot Semantic Segmentation Methods} \\
        \midrule
        SSP \cite{fan2022self} & Res-50 & ECCV-22 & 40.00 & 48.68 & 35.49 & 45.86 & 74.44 & 74.26 & 78.91 & 80.59 & 57.21 & 62.35 \\
        FPTrans \cite{zhang2022feature} & ViT-base & NIPS-22 & 38.36 & 49.30 & 48.65 & 60.37 & 80.92 & 82.91 & 80.74 & 83.65 & 62.17 & 69.06 \\
        HDMNet \cite{peng2023hierarchical} & Res-50 & CVPR-23 & 25.40 & 39.10 & 33.00 & 35.00 & 30.60 & 31.30 & 75.10 & 78.60 & 41.00 & 46.00 \\
        PerSAM \cite{zhang2023personalize}  & ViT-base & ICLR-24 & 36.08 & 40.65 & 23.27 & 25.33 & 29.95 & 30.05 & 60.92 & 66.53 & 37.56 & 40.64 \\
        VRP-SAM \cite{sun2024vrp} & ViT-base & CVPR-24 & 40.43 & 44.75 & 28.21 & 31.96 & 30.54 & 29.99 & 80.78 & 83.18 & 44.99 & 47.47 \\
        \midrule
        \multicolumn{13}{c}{Cross-domain Few-shot Semantic Segmentation Methods} \\
        \midrule
        PATNet$^{*}$ \cite{lei2022cross} & Res-50 & ECCV-22 & 37.89 & 42.97 & 41.16 & 53.58 & 66.61 & 70.20 & 78.59 & 81.23 & 56.06 & 61.99 \\
        ABCDFSS$^{*}$ \cite{herzog2024adapt} & Res-50 & CVPR-24 & 42.60 & 49.00 & 45.70 & 53.30 & 79.80 & 81.40 & 74.60 & 76.20 & 60.67 & 64.97 \\
        DRA \cite{su2024domain} & Res-50 & CVPR-24 & 41.29 & 50.12 & 40.77 & 48.87 & 82.35 & 82.31 & 79.05 & 80.40 & 60.86 & 65.42 \\
        APSeg \cite{he2024apseg} & ViT-base & CVPR-24 & 35.94 & 39.98 & 45.43 & 53.98 & 84.10 & 84.50 & 79.71 & 81.90 & 61.30 & 65.09 \\
        APM$^{*}$ \cite{tong2024lightweight} & Res-50 & NIPS-24 & 40.86 & 44.92 & 41.71 & 51.16 & 78.25 & 82.81 & 79.29 & 81.83 & 60.03 & 65.18 \\
        LoEC \cite{liu2025devil} & Res-50 & CVPR-25 & 44.10 & 49.67 & 38.21 & 47.04 & 81.02 & 82.73 & 78.51 & 80.60 & 60.46 & 65.01 \\
        LoEC \cite{liu2025devil} & ViT-base & CVPR-25 & 42.12 & 51.48 & \underline{52.91} & \underline{62.43} & 83.94 & 84.12 & \underline{81.05} & \underline{83.69} & \underline{65.01} & \underline{70.43} \\
        SDRC$^{*}$ \cite{tong2025self} & ViT-base & ICML-25 & 43.15 & 46.83 & 46.57 & 55.02 & 82.86 & 84.79 & 80.31 & 82.55 & 63.22 & 67.30 \\
        \midrule
        \textbf{HSL(Ours)}  & Res-50 & Ours & \textbf{46.13} & \underline{53.80} & 48.01 & 55.56 & \underline{84.57} & \underline{85.34} & 78.22 & 80.36 & 64.23 & 68.77 \\
        \textbf{HSL(Ours)} & ViT-base & Ours & \underline{45.77} & \textbf{54.56} & \textbf{59.36} & \textbf{64.62} & \textbf{85.95} & \textbf{86.25} & \textbf{81.89} & \textbf{83.84} & \textbf{68.24} & \textbf{72.32} \\
        \bottomrule
    \end{tabular}}
    \caption{Mean-IoU of 1-shot and 5-shot results compared with previous FSS and CD-FSS methods. The best and second-best methods are highlighted in \textbf{bold} and \underline{underlined}, respectively. $*$ indicates fine-tuning with target domain data.}
    \label{tab:performance}
\end{table*}

\section{Experiments}
\subsection{Datasets and Metric}
Following the settings in PATNet \cite{lei2022cross}, we train our model on the PASCAL VOC 2012 \cite{everingham2010pascal} dataset augmented with SBD \cite{hariharan2011semantic}, then evaluate the trained model on four target domain datasets: Deepglobe \cite{demir2018deepglobe}, ISIC \cite{codella2019skin, tschandl2018ham10000}, Chest X-ray \cite{candemir2013lung, jaeger2013automatic}, and FSS-1000 \cite{li2020fss}. Please refer to the appendix for more details.

We use mean Intersection over Union (mIoU) as the evaluation metric, and report average results over 5 random seeds. Each run consists of 1200 episodes sampled from Deepglobe, ISIC, and Chest X-ray respectively, and 2400 episodes from FSS-1000.

\begin{figure}[t]
\centering
    \includegraphics[width=0.99\linewidth]{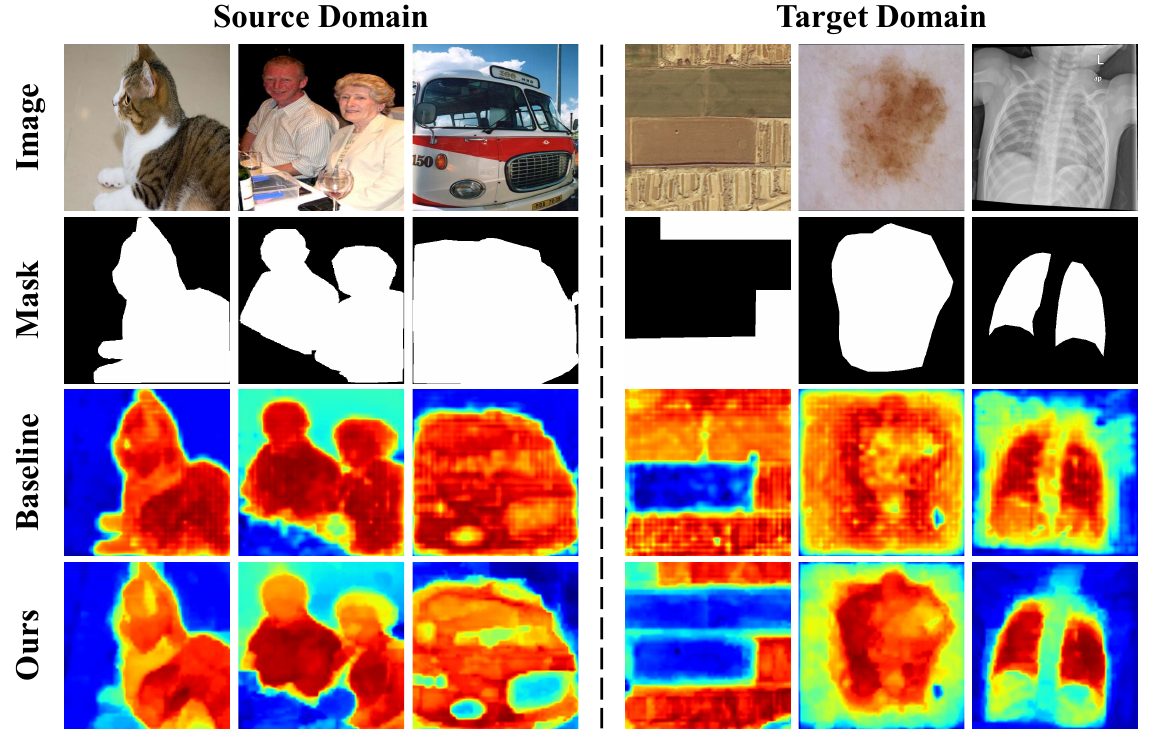}
    \caption{The heatmaps of foreground similarity maps for the source domain and target domains demonstrate that our method can extract hierarchical semantic features.}
    \label{fig:sim_map0}
\end{figure}

\subsection{Implementation Details}
Following \cite{liu2025devil}, we use ResNet-50 \cite{he2016deep} and ViT-B/16 \cite{dosovitskiy2020image} as our backbones. For the ResNet-50 backbone, we follow SSP \cite{fan2022self} by discarding the last backbone stage for better generalization, while resizing all images to $400 \times 400$ as per the default practice of previous CD-FSS methods and extracting low-level features at the end of stage-0. For the ViT-B/16 backbone, we follow FPTrans \cite{zhang2022feature} by resizing all images to $480 \times 480$, while extracting low-level features after patch embedding. All images are augmented following FPTrans during training. We use SGD to optimize our model for 5 epochs, with a momentum of 0.9, a weight decay of 5e-4, and a constant learning rate of 1e-3. All experiments are implemented on two NVIDIA GeForce RTX 4090 GPUs, with a batch size of 8 for 1-shot and 6 for 5-shot settings. We set the number of scales $L$ for superpixel masks to 4, with the corresponding number of superpixels (local regions) at each scale being $\{5^2,10^2,15^2,20^2\}$. The standard deviation $\sigma_{f}$ in the DSR is set to 0.25, and the standard deviations $\sigma_{g}$ are set to 0.1 for the ResNet-50 backbone and 0.6 for the ViT-B/16 backbone. Hyperparameters $K$, $\alpha$, $\beta$, and $\gamma$ are set to 9, 0.2, 40.0, and 0.1, respectively.

\begin{table}[!t]
    \centering
    \begin{tabular}{ccc|cc}
        \toprule
        DSR & HSM & PCMT & Res-50 & ViT-base\\
        \hline \rule{0pt}{2.5ex}
        $\times$ & $\times$ & $\times$ & 57.82 & 62.24 \\
        \checkmark & $\times$ & $\times$ & 60.44 & 64.55 \\
        $\times$ & \checkmark & $\times$ & 60.92 & 65.29 \\
        \checkmark & \checkmark & $\times$ & 62.97 & 67.05 \\
        \checkmark & \checkmark & \checkmark & \textbf{64.23} & \textbf{68.24} \\ 
        \bottomrule
    \end{tabular}
    \caption{Effects of each proposed component.}
    \label{tab:component}
\end{table}

\subsection{Comparison with State-of-the-art Methods}
In Tab. \ref{tab:performance}, we compare our method with existing FSS and CD-FSS approaches, including CNN-based and ViT-based methods. The results demonstrate significant improvements achieved by our method across different backbones under both 1-shot and 5-shot settings. Specifically, with the ResNet-50 backbone, our method surpasses the state-of-the-art DRA \cite{su2024domain} by 3.37$\%$ and 3.35$\%$ under 1-shot and 5-shot settings, respectively. With the ViT-B/16 backbone, our method surpasses the state-of-the-art LoEC \cite{liu2025devil} by 3.23$\%$ and 1.89$\%$ under 1-shot and 5-shot settings. Furthermore, our method outperforms existing methods across all four target domain datasets. These results clearly demonstrate the effectiveness of our method.

\subsection{Ablation Studies and Visualizations}
\subsubsection{Effects of each component.}
As shown in Tab. \ref{tab:component}, we conduct ablation studies on each proposed component. The first row represents our CNN-based and ViT-based baselines. DSR enhances the model's generalization ability by simulating data with diverse foreground-background style differences and overall style variations, improving the average mIoU of the CNN-based baseline by 2.62$\%$ and the ViT-based baseline by 2.31$\%$ under 1-shot setting. HSM guides the model to mine intra-class consistency and inter-class distinction at different granularities, respectively improving performance by 3.10$\%$ and 3.05$\%$ for the CNN-based and ViT-based baselines. PCMT introduces prototype confidence-weighted adaptive thresholds to mitigate segmentation ambiguity, further improving 
the performance.

\subsubsection{Effects of multi-scale superpixel masks.}
\begin{table}[!t]
    \centering
    \begin{tabular}{cccc|c}
        \toprule
        $5\times5$ & $10\times10$ & $15\times15$ & $20\times20$ & mean-IoU\\
        \hline \rule{0pt}{2.5ex}
        \checkmark & \checkmark & \checkmark & \checkmark & \textbf{67.05} \\
        $\times$ & \checkmark & \checkmark & \checkmark & 66.34 \\
        $\times$ & $\times$ & \checkmark & \checkmark & 65.84 \\
        $\times$ & $\times$ & $\times$ & \checkmark & 65.43 \\
        \bottomrule
    \end{tabular}
    \caption{Effects of multi-scale superpixel masks.}
    \label{tab:superpixel_size}
\end{table}
We use ViT-B/16 as the backbone and conduct ablation studies under 1-shot setting to validate the necessity of multi-scale superpixel masks. Specifically, we use four scales of superpixel masks in HSM with the number of regions being $\{5^2,10^2,15^2,20^2\}$ and sequentially remove them from coarse to fine. As shown in Tab. \ref{tab:superpixel_size}, the best performance is achieved when using all four scales, and each removal results in performance degradation. These results demonstrate that using multi-scale superpixel masks facilitates the model to capture category relationships at different granularities, thereby better adapting to target domains.

\subsubsection{Effects of different threshold computation strategies.}
\begin{table}[!t]
    \setlength{\tabcolsep}{4.5pt}
    \centering
        \begin{tabular}{c|ccccc}
            \toprule
            \small{Strategy} & \small{Deepglobe} & \small{ISIC} & \small{Chest} & \small{FSS} & \small{Average}\\
            \hline \rule{0pt}{2.5ex}
            \small{Fixed at 0} & \small{44.54} & \small{55.79} & \small{85.93} & \small{\textbf{81.93}} & \small{67.05}\\
            \small{OTSU} & \small{45.57} & \small{\textbf{59.98}} & \small{85.81} & \small{80.43} & \small{67.95}\\
            \small{PCMT (Ours)} & \small{\textbf{45.77}} & \small{59.36} & \small{\textbf{85.95}} & \small{81.89} & \small{\textbf{68.24}}\\
            \bottomrule
        \end{tabular}
    \caption{Effects of threshold computation strategies.}
    \label{tab:threshold}
\end{table}
To validate the effects of our PCMT, we use ViT-B/16 as the backbone and compare the performance of different thresholding strategies under the 1-shot setting, with the results shown in Tab. \ref{tab:threshold}. The adaptive threshold calculated by OTSU \cite{otsu1975threshold} demonstrates significant performance improvement on the ISIC dataset, which is prone to segmentation ambiguity. However, it shows a notable performance drop on the FSS-1000 dataset, which is close to the source domain. When the threshold is fixed at 0, the opposite result is observed. Our PCMT dynamically adjusts thresholds based on the prototype confidence of each sample, achieving a better balance across different target domain data.

\begin{figure}[t]
\centering
    \includegraphics[width=0.99\linewidth]{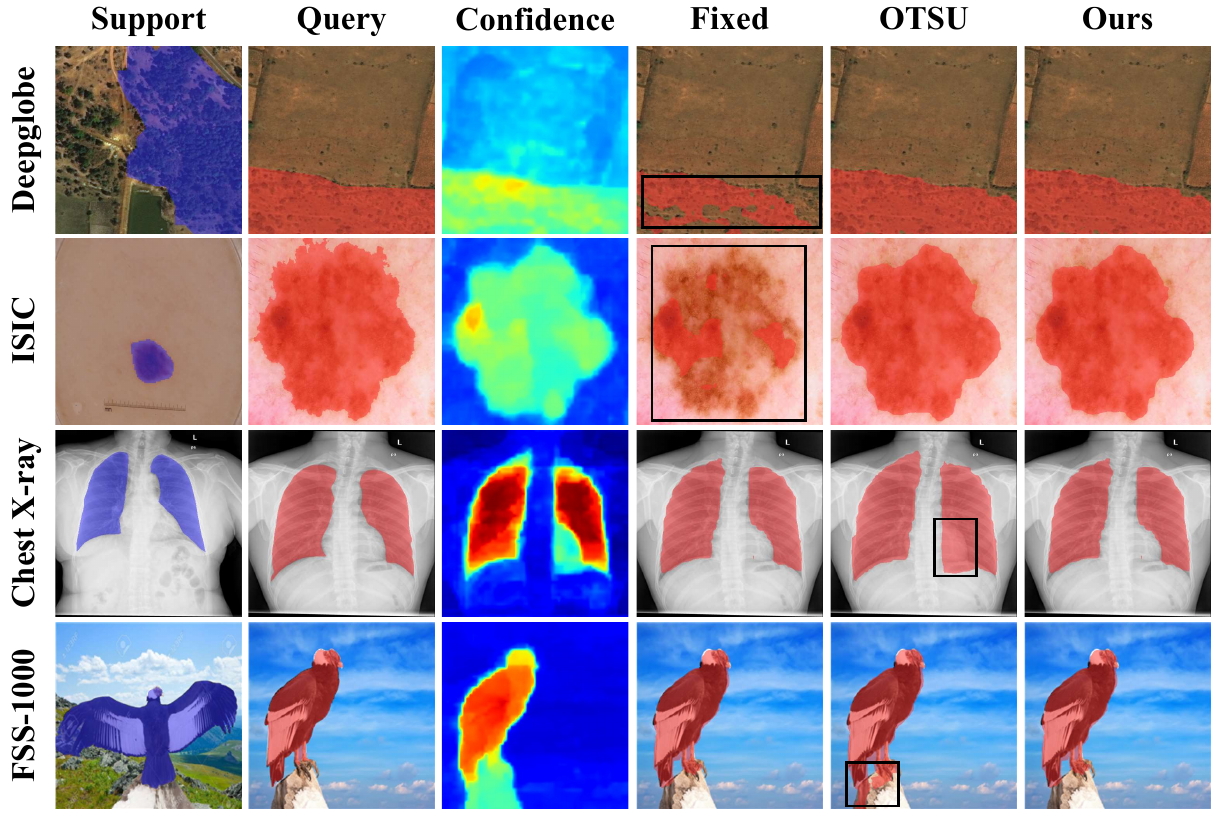}
    \caption{The heatmaps of foreground confidence maps and segmentation results using different thresholding methods.}
    \label{fig:conf_map}
\end{figure}

\subsubsection{Visualizations.}
We use ground truth masks to extract foreground prototypes and calculate similarity maps, with the corresponding heatmaps visualized in Fig. \ref{fig:sim_map0}. Our method extracts more hierarchical features from the source domain, enabling discrimination of semantics at varying granularities in target domains. Additionally, we visualize the heatmaps of the foreground confidence maps and the segmentation results using different thresholding methods, as shown in Fig. \ref{fig:conf_map}. For cases with segmentation ambiguity (top two rows), our method effectively alleviates these ambiguities. For cases without segmentation ambiguity (bottom two rows), our method avoids introducing additional errors at the boundaries. This demonstrates the flexibility and effectiveness of our method.
We also present extensive ablation studies and more visualizations in the appendix.

\section{Conclusion}
In this paper, we propose an HSL framework to address both inter-domain style gaps and segmentation granularity gaps in CD-FSS. Specifically, we propose a DSR module that sequentially performs foreground style randomization and global style randomization on training data, simulating target domain data with diverse foreground-background style differences and overall style variations. Additionally, we propose an HSM module that leverages multi-scale superpixel masks to extract multi-scale region prototypes to enhance features. DSR and HSM facilitate the model in extracting hierarchical semantic features, enabling better generalization to target domains. Furthermore, we propose a PCMT module to mitigate segmentation ambiguity, further improving segmentation performance. Extensive experiments show that our method achieves state-of-the-art performance on four datasets with different domain gaps.

\section*{Acknowledgements}
This work was supported in part by the National Natural Science Foundation of China under the Grants No. 62371235 and No. U25A20444, in part by the Key Research and Development Plan of Jiangsu Province under Grant No. BE2023008-2.

\bibliography{aaai2026}

\end{document}


\maketitle
\section{Multi-scale Superpixel Masks Generation}
Superpixel segmentation divides an image into multiple local regions, where each region consists of adjacent pixels with similar features or identical semantics. Our insight is that superpixel segmentation can create local regions of various sizes, approximating semantic segmentation results at multiple granularities. Therefore, we generate multi-scale superpixel masks to guide the model in learning hierarchical semantic features.

Specifically, given an original image $\mathbf{I}\in\mathbb{R}^{3\times H\times W}$ and a specified superpixel number $\sqrt{n_i}\times \sqrt{n_i}$, we first downsample the image to obtain $\mathbf{I}^{\prime}\in\mathbb{R}^{3\times d\sqrt{n_i}\times d\sqrt{n_i}}$, where $d$ denotes the downsampling scale during feature extraction by the image encoder. In deep learning-based superpixel segmentation methods, $d$ is typically set to 16. Subsequently, we uniformly divide $\mathbf{I}^{\prime}$ into $\sqrt{n_i}\times \sqrt{n_i}$ initial superpixel grids and construct a neighborhood relation map $\mathbf{Z}^{sp}\in \mathbb{R}^{9\times d\sqrt{n_i}\times d\sqrt{n_i}}$ that stores the indexes of the nine superpixels adjacent to each pixel.

Deep learning-based methods formulate superpixel segmentation as a dense classification task, predicting the probability of each pixel belonging to adjacent superpixels. This process is formulated as:
\begin{equation}
\mathbf{Q}^{sp}=\operatorname{Dec}(\operatorname{Enc}(\mathbf{I}^{\prime})),
\end{equation}
where $\mathbf{Q}^{sp}\in\mathbb{R}^{9\times d\sqrt{n_i}\times d\sqrt{n_i}}$ denotes the neighborhood probability map, representing the probability of each pixel belonging to the nine adjacent superpixels. $\operatorname{Enc}$ and $\operatorname{Dec}$ represent the image encoder and decoder, respectively. Based on $\mathbf{Q}^{sp}$ and $\mathbf{Z}^{sp}$, we assign each pixel to the superpixel with the highest probability, obtaining the superpixel mask $\mathbf{M}$:
\begin{equation}
\mathbf{M}(x,y)=\mathbf{Z}^{sp}(\arg\max_{c}{\mathbf{Q}^{sp}(c,x,y)},x,y),
\end{equation}
where $c$ denotes the channel index and $(x,y)$ represents spatial coordinates.

We use \cite{xu2024learning} to implement the above process, which can also be seamlessly replaced with other deep learning-based superpixel segmentation methods \cite{yang2020superpixel, wang2021ainet, yuan2021sin}. By sequentially setting the superpixel number $n_i$ to $L$ different values, we obtain multi-scale superpixel masks $\mathcal{M}^{sp}=\{\mathbf{M}_i\}_{i=1}^L$.

\begin{figure}[t]
    \centering
    \includegraphics[width=\linewidth]{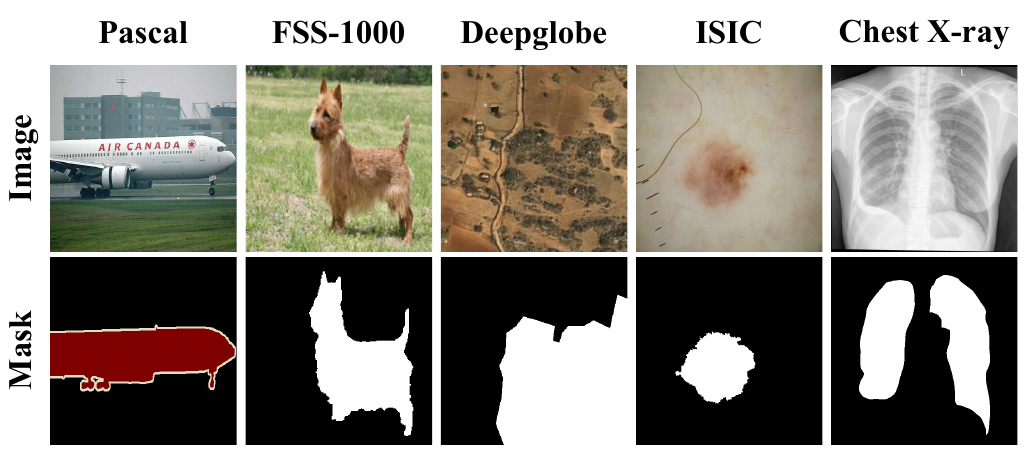}
    \caption{Images and their corresponding ground truth masks sampled from the source domain dataset and four target domain datasets.}
    \label{fig:samples}
\end{figure}

\section{Loss Function}
We use Binary Cross Entropy (BCE) loss to train our model:
\begin{equation}
\mathcal{L}_{final}=\operatorname{BCE}(\operatorname{cos}(\hat{\mathbf{p}}^{fg},\hat{\mathbf{F}}_q),\operatorname{cos}(\hat{\mathbf{p}}^{bg},\hat{\mathbf{F}}_q),\mathbf{M}_q),
\end{equation}
where $\hat{\mathbf{p}}^{fg}$ and $\hat{\mathbf{p}}^{bg}$ denote fused prototypes computed using hierarchical semantic features $\hat{\mathbf{F}}$.

To better supervise the training of the image encoder, we also introduce the same loss function as SSP \cite{fan2022self}:
\begin{equation}
\begin{split}
\mathcal{L}_{ssp}=&\operatorname{BCE}(\operatorname{cos}(\mathbf{p}^{fg},\mathbf{F}_q^h),\operatorname{cos}(\mathbf{p}^{bg},\mathbf{F}_q^h),\mathbf{M}_q)\\+&\operatorname{BCE}(\operatorname{cos}(\mathbf{p}_q^{fg},\mathbf{F}_q^h),\operatorname{cos}(\mathbf{p}_q^{bg},\mathbf{F}_q^h),\mathbf{M}_q)\\+&\lambda \operatorname{BCE}(\operatorname{cos}(\mathbf{p}_s^{fg},\mathbf{F}_s^h),\operatorname{cos}(\mathbf{p}_s^{bg},\mathbf{F}_s^h),\mathbf{M}_s),
\end{split}
\end{equation}
where $\mathbf{p}^{*}$, $\mathbf{p}_q^{*}$, and $\mathbf{p}_s^{*}$ represent fused prototypes, query prototypes, and support prototypes computed using high-level features $\mathbf{F}^h$, respectively. $\lambda=0.2$ is the weight parameter.

Finally, the total loss function is $\mathcal{L}=\mathcal{L}_{final}+\mathcal{L}_{ssp}$.

\begin{table*}[t]
    \setlength{\tabcolsep}{3pt}
    \centering
    \begin{tabular}{c|c|c|cc|cc|cc|cc|cc}
        \toprule
        \rule{0pt}{2.5ex}
        \multirow{2}*{Methods} & \multirow{2}*{Backbone} & \multirow{2}*{Mark} & \multicolumn{2}{c|}{Deepglobe} & \multicolumn{2}{c|}{ISIC} & \multicolumn{2}{c|}{Chest X-ray} & \multicolumn{2}{c|}{FSS-1000} & \multicolumn{2}{c}{Average} \\
        \cline{4-13}
        \rule{0pt}{2.5ex}
        & & & \multicolumn{1}{c|}{1-shot} & 5-shot & \multicolumn{1}{c|}{1-shot} & 5-shot & \multicolumn{1}{c|}{1-shot} & 5-shot & \multicolumn{1}{c|}{1-shot} & 5-shot & \multicolumn{1}{c|}{1-shot} & 5-shot \\
        \midrule
        IFA \shortcite{nie2024cross} & Res-50 & CVPR-24 & 50.6 & 58.8 & 66.3 & 69.8 & 74.0 & 74.6 & 80.1 & 82.4 & 67.8 & 71.4 \\
        GPRN \shortcite{peng2025sam} & Res-50 + ViT-base & AAAI-25 & \underline{51.7} & 59.3 & 66.8 & 72.2 & 87.0 & 87.1 & 81.1 & 82.6 & 71.7 & 75.3 \\
        SDRC \shortcite{tong2025self} & ViT-base & ICML-25 & 51.1 & \underline{59.4} & \underline{69.7} & 72.5 & 84.1 & 87.2 & \underline{83.1} & \underline{85.7} & 72.0 & 76.2 \\
        \textbf{HSL(Ours)}  & Res-50 & Ours & 50.6 & 59.0 & 69.2 & \underline{75.9} & \underline{87.1} & \underline{87.3} & 82.4 & 84.4 & \underline{72.3} & \underline{76.7} \\
        \textbf{HSL(Ours)} & ViT-base & Ours & \textbf{51.8} & \textbf{59.5} & \textbf{75.5} & \textbf{78.4} & \textbf{88.0} & \textbf{88.5} & \textbf{85.5} & \textbf{86.2} & \textbf{75.2} & \textbf{78.1} \\
        \bottomrule
    \end{tabular}
    \caption{Mean-IoU of 1-shot and 5-shot results compared with previous CD-FSS methods under the setting of IFA \cite{nie2024cross}. The best and second-best methods are highlighted in \textbf{bold} and \underline{underlined}, respectively.}
    \label{tab:performance_ifa}
\end{table*}

\section{Dataset Details}
Following the settings in PATNet \cite{lei2022cross}, we train our model on the PASCAL VOC 2012 \cite{everingham2010pascal} dataset augmented with SBD \cite{hariharan2011semantic}, then evaluate the trained model on four target domain datasets: Deepglobe \cite{demir2018deepglobe}, ISIC \cite{codella2019skin, tschandl2018ham10000}, Chest X-ray \cite{candemir2013lung, jaeger2013automatic}, and FSS-1000 \cite{li2020fss}. The images sampled from each dataset are illustrated in Fig. \ref{fig:samples}.

\textbf{Deepglobe} is a satellite remote sensing dataset, where each image provides pixel-level dense annotations for 7 classes: areas of urban, agriculture, rangeland, forest, water, barren, and unknown. Following the preprocessing strategy of PATNet \cite{lei2022cross}, we crop the original images with a resolution of $2448 \times 2448$ into 6 pieces. After filtering single-class images and the `unknown' class, we finally obtain 5666 images with a resolution of $408 \times 408$ for evaluation.

\textbf{ISIC} is a medical image dataset for skin cancer screening, containing 2596 skin lesion images including three major types of skin lesions. The original image resolution is approximately $1022 \times 767$, and all images are uniformly downsampled to $512 \times 512$.

\textbf{Chest X-ray} is a dataset of tuberculosis diagnosis X-ray images, containing 566 medical images collected from 58 tuberculosis-positive cases and 80 normal cases. The original image resolution is $4020 \times 4892$, and all images are uniformly downsampled to $1024 \times 1024$.

\textbf{FSS-1000} is a natural image dataset designed for few-shot segmentation, containing 1000 target categories, with 10 images annotated per category. We evaluate the trained model on the official test split, which consists of 240 categories and a total of 2400 images.

\section{Comparison with Methods under Special Setting}
Some CD-FSS methods follow the setting of IFA \cite{nie2024cross} and treat all samples in a batch as the same class when calculating mean-IoU, with a batch size of 96. This often leads to results that are higher than the true mean-IoU. Additionally, this setting includes a fine-tuning process with 20 epochs, in which one or five samples are randomly selected from each class for fine-tuning in each epoch. For a fair comparison, we conduct experiments under the setting of IFA and report results in Tab. \ref{tab:performance_ifa}. The results demonstrate that our method also achieves state-of-the-art performance on all four target domain datasets under this setting.

\section{Model Efficiency}
\begin{table}[htb]
    \centering
    \setlength{\tabcolsep}{3.2pt}
        \begin{tabular}{c|ccccc}
            \toprule
            Method & Params (M) & FLOPs (G) & FPS & mean-IoU\\
            \hline \rule{0pt}{2.5ex}
            DRA \shortcite{su2024domain} & 59.3 & 257.0 & 36.90 & 60.86 \\
            Ours & 17.2 & 226.9 & 43.29 & 64.23 \\
            \bottomrule
        \end{tabular}
    \caption{Model efficiency and performance under 1-shot setting, including the trainable parameter number, computational complexity (FLOPs), inference speed (FPS), and mean-IoU.}
    \label{tab:efficiency}
\end{table}

We compare the model efficiency of our method with the current CNN-based state-of-the-art DRA \cite{su2024domain} under 1-shot setting. This comparison includes the trainable parameter number, computational complexity (FLOPs), and inference speed (FPS), with all experiments conducted on an NVIDIA GeForce RTX 4090 GPU. As shown in Tab. \ref{tab:efficiency}, our method demonstrates both greater model efficiency and higher segmentation performance. In practice, we can also extract support features and compute support prototypes in advance, further reducing computational complexity and inference time by approximately half.

\section{Ablation Studies}
We provide more method comparison and hyperparameter analysis through extensive experiments. Unless otherwise specified, all experiments employ ViT-base as backbone and are conducted under 1-shot setting.

\subsubsection{Effects of each style randomization module.}
\begin{table}[htb]
    \centering
    \begin{tabular}{cc|cc}
        \toprule
        Foreground & Global & Res-50 & ViT-base\\
        \hline \rule{0pt}{2.5ex}
        $\times$ & $\times$ & 57.82 & 62.24 \\
        \checkmark & $\times$ & 58.71 & 62.92 \\
        $\times$ & \checkmark & 59.81 & 63.75 \\
        \checkmark & \checkmark & \textbf{60.44} & \textbf{64.55} \\
        \bottomrule
    \end{tabular}
    \caption{Ablation studies for the effects of each style randomization module.}
    \label{tab:randomization}
\end{table}
To explore the effects of foreground and global style randomization on model performance, we conduct additional ablation studies, with the results shown in Tab. \ref{tab:randomization}. The results indicate a significant decline in model performance when neither foreground style randomization nor global style randomization is used. This demonstrates the existence of inter-domain variations in foreground-background differences and overall styles, while both randomization processes partially mitigate these domain gaps.

\subsubsection{Effects of different $\sigma_{g}$.}
\begin{figure}[htb]
    \centering
    \includegraphics[width=\linewidth]{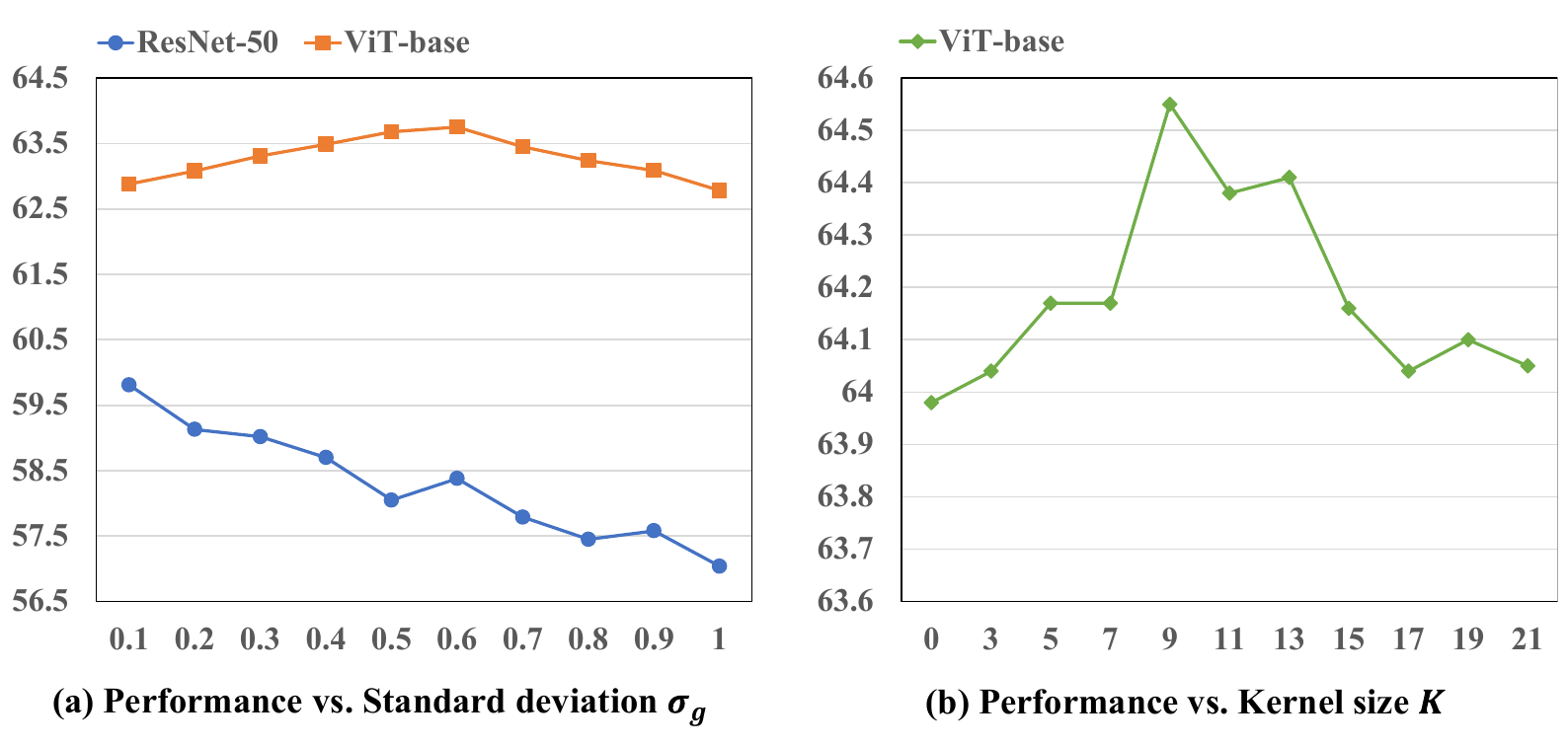}
    \caption{(a) The mIoU curves corresponding to different sampling standard deviations $\sigma_{g}$ in global style randomization. (b) The mIoU curves corresponding to different kernel sizes $K$ in foreground style randomization.}
    \label{fig:line_chart}
\end{figure}
To validate the effects of varying levels of randomization on model performance, we set different sampling standard deviations $\sigma_{g}$ for random convolution parameters in global style randomization, as shown in Fig. \ref{fig:line_chart}(a). When $\sigma_{g}$ is set to 0.1 and 0.6, respectively, methods using ResNet-50 and ViT-base as backbones achieve the best performance, which indicates that ViT-base is more prone to overfitting to source domain styles, requiring a higher degree of style randomization to alleviate this issue. However, larger $\sigma_{g}$ is not always better: excessively large $\sigma_{g}$ leads to training instability, which consequently degrades model performance.

\subsubsection{Effects of sampling distributions of $\omega$.}
\begin{table}[htb]
    \centering
    \begin{tabular}{c|ccc}
        \toprule
        Distribution & $N(0, 0.25)$ & $U(-0.5, 0.5)$ & $U(0, 0.5)$\\
        \hline \rule{0pt}{2.5ex}
        mean-IoU & \textbf{64.55} & 64.37 & 63.92 \\
        \bottomrule
    \end{tabular}
    \caption{Ablation studies for the effects of sampling distributions of $\omega$.}
    \label{tab:distributions}
\end{table}

In foreground style randomization, we select Gaussian distribution to generate the weights of local styles, allowing the foreground region’s style to approach or diverge from the local region’s style to varying degrees. As shown in Tab. \ref{tab:distributions}, our method is not limited to the Gaussian distribution and achieves comparable performance when using other sampling distributions. However, the model exhibits the lowest performance when the sampling distribution is $U(0,0.5)$, demonstrating the necessity for foreground styles to deviate from local styles, which can simulate more diverse target domain data.

\subsubsection{Effects of different kernel size $K$.}
We set different values for the kernel size $K$ of max pooling and average pooling in foreground style randomization to explore its effects on results, as shown in Fig. \ref{fig:line_chart}(b). $K=0$ indicates no preprocessing on the mask, which results in visible foreground-background boundaries in the randomized image and may lead to overfitting. When $K$ is set to 9, the model achieves the best performance.

\subsubsection{Effects of different low-level feature fusion strategies.}
\begin{table}[htb]
    \centering
    \setlength{\tabcolsep}{3.2pt}
        \begin{tabular}{c|ccccc}
            \toprule
            Strategy & Deepglobe & ISIC & Chest & FSS & Average\\
            \hline \rule{0pt}{2.5ex}
            Baseline & 43.73 & 55.54 & 84.01 & 81.78 & 66.27 \\
            Original $F^l$ & 43.55 & 54.09 & 83.14 & 81.74 & 65.63 \\
            Enhanced $F^l$ & \textbf{44.54} & \textbf{55.79} & \textbf{85.93} & \textbf{81.93} & \textbf{67.05} \\
            \bottomrule
        \end{tabular}
    \caption{Ablation studies for the effects of different low-level feature fusion strategies.}
    \label{tab:fusion}
\end{table}
To explore the effects of low-level features in HSM, we compare the performance of different low-level feature fusion strategies, and the results are shown in Tab. \ref{tab:fusion}. First, we use the strategy without low-level features as the baseline. When the original low-level features are directly fused with high-level features, the performance decreases instead, indicating that original low-level features may disrupt the semantic information in high-level features. When the low-level features are enhanced before being fused, the model performance improves, indicating that the enhanced low-level features provide additional detailed information to the high-level features.

\subsubsection{Effects of different $\alpha$.}
\begin{table}[htb]
    \centering
        \begin{tabular}{c|ccccc}
            \toprule
            $\alpha$ & Deepglobe & ISIC & Chest & FSS & Average\\
            \hline \rule{0pt}{2.5ex}
            0.0 & 43.73 & 55.54 & 84.01 & 81.78 & 66.27 \\
            0.2 & \textbf{44.54} & 55.79 & \textbf{85.93} & \textbf{81.93} & \textbf{67.05} \\
            0.4 & 43.87 & \textbf{56.01} & 84.62 & 81.74 & 66.56 \\
            0.6 & 44.25 & 54.45 & 82.79 & 80.92 & 65.60 \\
            \bottomrule
        \end{tabular}
    \caption{Ablation studies for the effects of different $\alpha$.}
    \label{tab:alpha}
\end{table}
We set the fusion weight $\alpha$ of low-level features to 0.0, 0.2, 0.4, and 0.6 to investigate its impact on model performance. As shown in Tab. \ref{tab:alpha}, the model achieves the best performance when $\alpha$ is set to 0.2, indicating that a small number of low-level features can enhance the detail discrimination ability of high-level features. However, an excessive number of low-level features weakens the semantic information of high-level features.

\subsubsection{Effects of different superpixel segmentation methods.}
\begin{table}[htb]
    \centering
    \setlength{\tabcolsep}{3.2pt}
        \begin{tabular}{c|ccccc}
            \toprule
            Method & Deepglobe & ISIC & Chest & FSS & Average\\
            \hline \rule{0pt}{2.5ex}
            SCN \shortcite{yang2020superpixel} & 45.62 & \textbf{59.59} & 85.58 & 81.70 & 68.12 \\
            SIN \shortcite{yuan2021sin} & 45.50 & 58.65 & 85.35 & 81.63 & 67.78 \\
            AINet \shortcite{wang2021ainet} & 45.60 & 58.94 & 85.42 & 81.75 & 67.93 \\
            CDS \shortcite{xu2024learning} & \textbf{45.77} & 59.36 & \textbf{85.95} & \textbf{81.89} & \textbf{68.24} \\
            \bottomrule
        \end{tabular}
    \caption{Ablation studies for the effects of different superpixel segmentation methods.}
    \label{tab:superpixel_method}
\end{table}

We use different superpixel segmentation methods to extract multi-scale superpixel masks. As shown in Tab. \ref{tab:superpixel_method}, using different superpixel segmentation methods does not have significant effects on the results, demonstrating that our method is not sensitive to the quality of multi-scale superpixel masks.

\begin{figure*}[t!]
    \centering
    \includegraphics[width=\linewidth]{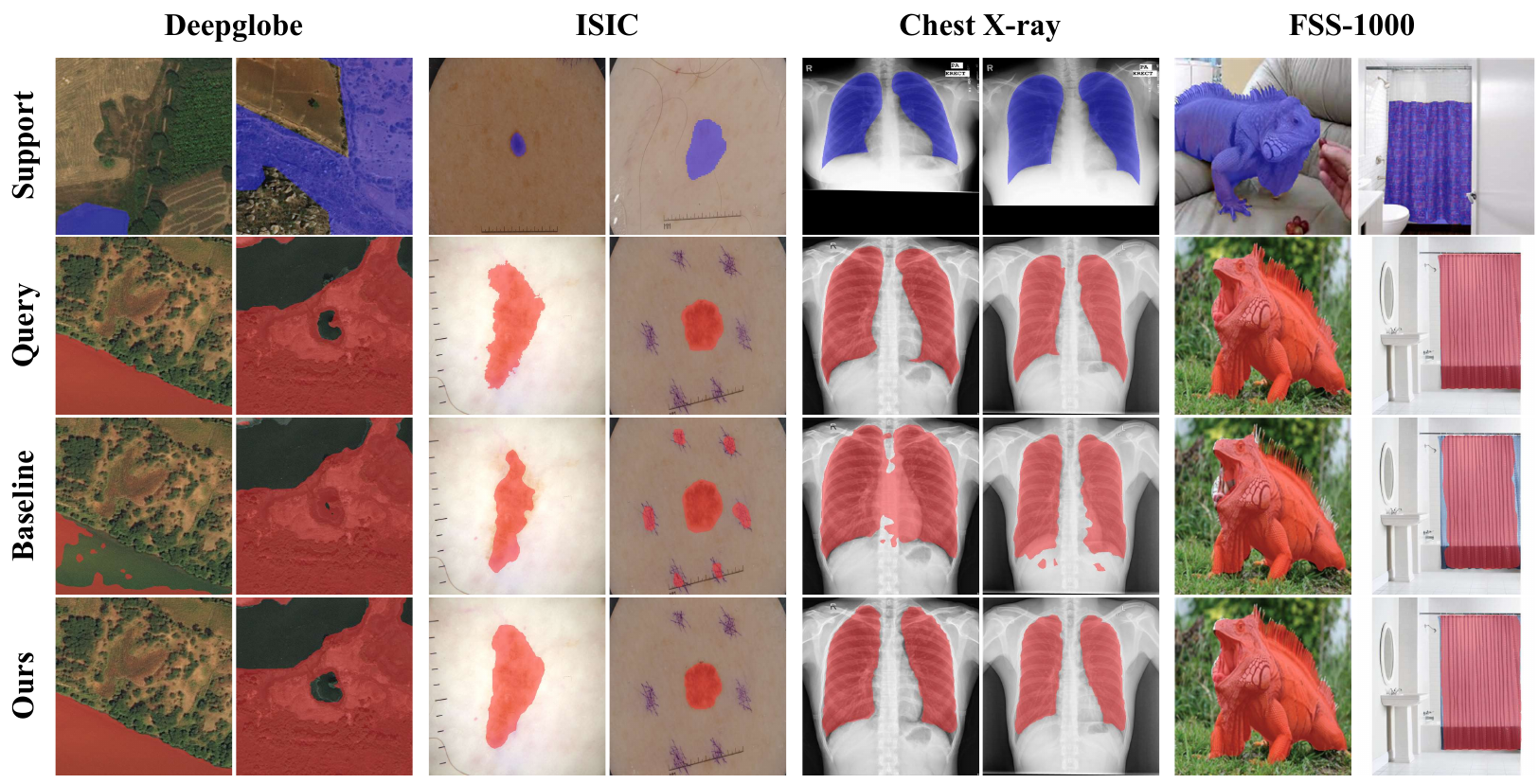}
    \caption{Qualitative results of our model for 1-way 1-shot setting. The blue parts represent support masks and the red parts represent query masks and query predictions.}
    \label{fig:results}
\end{figure*}

\begin{figure*}[t!]
\centering
    \includegraphics[width=0.97\linewidth]{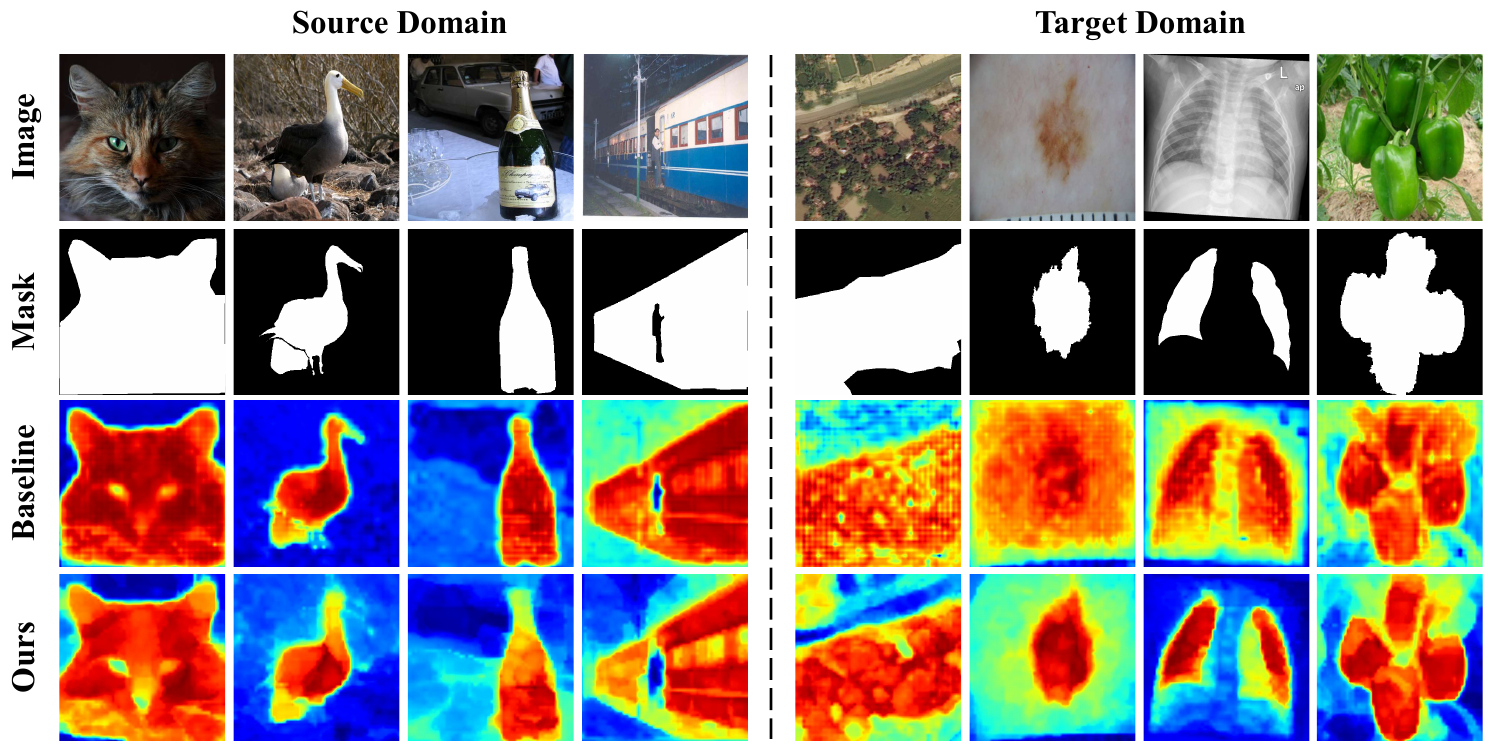}
    \caption{The heatmaps of foreground similarity maps for the source domain and target domains demonstrate that our method can extract hierarchical semantic features, effectively distinguishing the foreground and background of target domain images with different segmentation granularity gaps.}
    \label{fig:sim_map}
\end{figure*}

\section{More Visualizations}
\subsection{Qualitative Results}
In Fig. \ref{fig:results}, we present some additional qualitative results of our ViT-based method for 1-way 1-shot segmentation.

\subsection{Hierarchical Semantic Features}
To validate our model's ability to extract hierarchical semantic features, we utilize ground truth masks of images to extract foreground prototypes, then calculate similarity maps between the foreground prototypes and features, and visualize the corresponding heatmaps in Fig. \ref{fig:sim_map}. The visualizations of the source domain show that the features extracted by our model have different similarities across different regions within the foreground, while the features extracted by the baseline show consistent similarity across the entire foreground region, indicating that our model can extract more hierarchical semantic features. This ability to extract hierarchical semantic features allows our model to effectively distinguish the foreground and background of target domain images with different segmentation granularity gaps.

\bibliography{aaai2026}